%% file: main.tex
\documentclass[runningheads]{llncs}
\usepackage[T1]{fontenc}
\usepackage{graphicx}
\usepackage{xcolor}
\usepackage{amsmath}
\usepackage{amssymb}
\usepackage{textcomp}
\usepackage{amsfonts}
\usepackage{multirow}
\usepackage{tabularray}
\UseTblrLibrary{booktabs}
\usepackage{adjustbox}
\usepackage{array}
\usepackage{enumitem}
\usepackage{hyperref}

\NewDocumentCommand{\todo}
{ mO{} }{\textcolor{magenta}{\textsuperscript{\textit{TODO}}\textsf{\textbf{\small[#1]}}}}

\begin{document}

\title{FairDiff: Fair Segmentation with Point-Image Diffusion}

\author{Wenyi Li\textsuperscript{*} \inst{1} \and
Haoran Xu\textsuperscript{*} \inst{1} \and
Guiyu Zhang\textsuperscript{*} \inst{1} \and
Huan-ang Gao\inst{1} \and
Mingju Gao\inst{1} \and
Mengyu Wang\inst{2} \and
Hao Zhao\textsuperscript{\textdagger} \ \inst{1}
}


\authorrunning{W. Li et al.}
\titlerunning{FairDiff: Fair Segmentation with Point-Image Diffusion}

\institute{Institute for AI Industry Research (AIR), Tsinghua University \and
Harvard Ophthalmology AI Lab, Harvard University 
\email{liwenyi19@mails.ucas.ac.cn, zhaohao@air.tsinghua.edu.cn}\\
\setcounter{footnote}{1}
\footnotetext{\textsuperscript{*} Indicates Equal Contribution. \textsuperscript{\textdagger} Indicates Corresponding Author.}
}

\maketitle              

\begin{abstract}
Fairness is an important topic for medical image analysis, driven by the challenge of unbalanced training data among diverse target groups and the societal demand for equitable medical quality. In response to this issue, our research adopts a data-driven strategy—enhancing data balance by integrating synthetic images. However, in terms of generating synthetic images, previous works either lack paired labels or fail to precisely control the boundaries of synthetic images to be aligned with those labels. To address this, we formulate the problem in a joint optimization manner, in which three networks are optimized towards the goal of empirical risk minimization and fairness maximization. On the implementation side, our solution features an innovative Point-Image Diffusion architecture, which leverages 3D point clouds for improved control over mask boundaries through a point-mask-image synthesis pipeline. This method outperforms significantly existing techniques in synthesizing scanning laser ophthalmoscopy (SLO) fundus images. By combining synthetic data with real data during the training phase using a proposed Equal Scale approach, our model achieves superior fairness segmentation performance compared to the state-of-the-art fairness learning models. Code is available at \href{https://github.com/wenyi-li/FairDiff}{https://github.com/wenyi-li/FairDiff}.

\keywords{Fairness Learning \and  Image Synthesis \and Fundus Image Segmentation \and Diffusion Models}
\end{abstract}

\section{Introduction}

The pursuit of fairness in medical image analysis is an important topic because training data for different groups are usually unbalanced while the society calls for equitable medical quality across diverse target groups. 
To address the issue of unfairness, there are two principal strategies, optimization-driven and data-driven approaches. 
The optimization-driven approaches \cite{zemel2013learning,madras2018learning,sagawa2019distributionally,hardt2016equality,pleiss2017fairness} apply fair learning methods during training over perception models \cite{gao2023semi,tian2023unsupervised,gao2023dqs3d}, such as adjusting the weight of loss for different sensitive attributes or incorporating additional regularization losses to minimize bias across groups. 
By contrast, a more principled solution, what we name as data-driven approach, addresses the root cause of the issue—unbalanced data distribution through the augmentation of datasets with additional images from underrepresented groups. 
However, acquiring medical images, such as scanning laser ophthalmoscopy (SLO) fundus images, from vulnerable and under-represented populations proves challenging \cite{sharma2021improving,shepherd2020under}. Therefore, we resort to image synthesis methods \cite{rombach2022high,zhang2023adding,gao2024scp} to generate additional data, aiming to improve the fairness of results.

A recent study FairSeg \cite{tian2024fairseg} proposes the first fairness dataset for medical segmentation, including 10,000 SLO fundus images from 10,000 patients with pixel-wise disc and cup segmentation mask annotation. The cup-disc area is important for diagnosing a range of eye conditions. However, the anatomical structures of the fundus vary across different racial groups. For instance, Blacks often have a larger cup-to-disc ratio than other races, and Asians are more prone to angle-closure glaucoma than Whites. Therefore, the diversity of the cup-disc area present challenges to the synthesis of images. 

\begin{figure}[!t]
    \centering
    \includegraphics[width=0.8\linewidth]{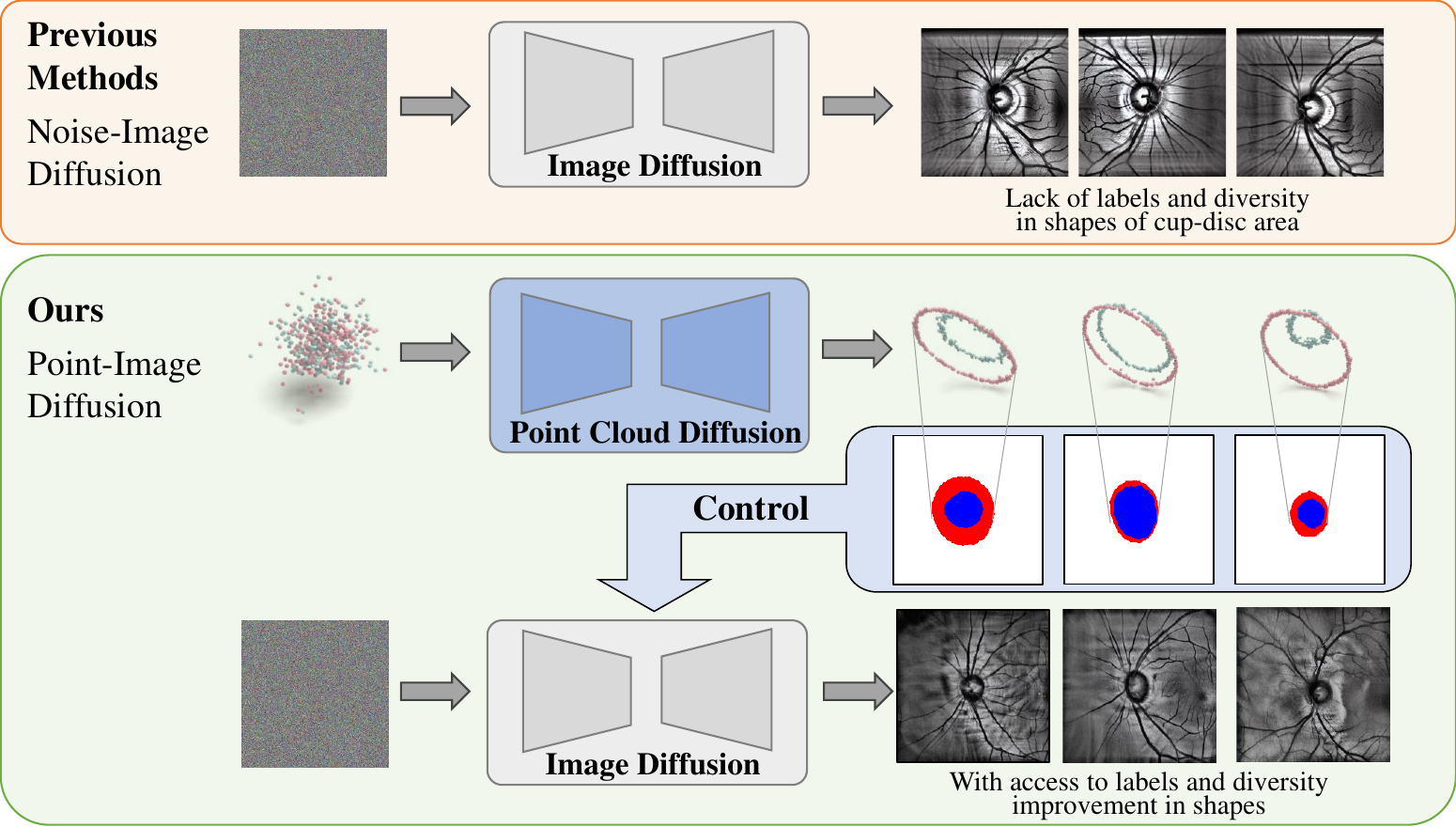}
    \caption{\textbf{Comparison of Traditional Noise-Image Diffusion and Our Point-Image Diffusion Methods.} We transform 2D mask data into a 3D point cloud format, leveraging the spatial coordinates to delineate boundaries more accurately.}
    \label{fig:teaser}
\end{figure}

Previous medical image synthesis works \cite{khader2022medical,friedrich2024wdm,dalmaz2022resvit} mainly focused on the direct generation of medical images. Although these synthesized images often closely mimic the distribution of real images, they lack paired labels, and the process of annotating these images is time-consuming and labor-intensive.
Several mask-to-image works, like Freemask \cite{yang2024freemask} and OASIS  \cite{sushko2020you}, can generate images from masks but utilize the same set of masks for both real and synthetic samples, which lead to a lack of diversity.
SEGGEN \cite{ye2023seggen} proposed MaskSyn which can generate masks through 2D diffusion model \cite{podell2023sdxl}. However, accurately controlling these mask boundaries in a two-dimensional space using GANs \cite{goodfellow2014generative,arjovsky2017wasserstein,karras2019style} and 2D diffusion models \cite{ho2020denoising,rombach2022high,dhariwal2021diffusion} is challenging, because of the inherent constraints in capturing boundaries with pixel-level detail \cite{shen2023boundary,shen2023measuring}.

To address these challenges, we explore the utilization of sampling points that convert pixel boundaries into spatial coordinates to enhance boundary control. Sampling points on a 2D mask results in a 2D point cloud. However, to distinguish different category boundaries, we transform the point cloud into a 3D format for better point feature learning, where the z-axis is used to mark categories, while the x and y coordinates retain their roles as the plane coordinates. The process is illustrated in Fig.~\ref{fig:teaser}.

In this paper, we formulate the fairness problem in a joint optimization manner, in which three networks are optimized towards the goal of empirical risk minimization and fairness maximization. On the implementation side, we introduce a novel Point-Image Diffusion architecture. In this framework, we first generate segmentation masks using point cloud diffusion and then synthesize images based on the control of these synthesized masks. After acquiring a substantial amount of synthetic data for various minority groups, we employ an equal-scale data-combining approach to ensure that the populations of each sensitive attribute group are balanced.

\textbf{Contributions.} In summary, our contributions are as follows: \textbf{(1)} We introduce a novel Point-Image diffusion approach for medical image synthesis, which utilizes 3D point clouds to enhance mask boundary control. This method outperforms significantly existing techniques in SLO fundus image synthesis. \textbf{(2)} Downstream segmentation tasks verified the efficacy of the synthesized data, demonstrating improvement in model segmentation performance. \textbf{(3)} By simply integrating synthetic and real data in the training phase, we achieve superior fairness segmentation performance compared to the state-of-the-art fairness learning models.

\begin{figure}[!t]
    \centering
    \includegraphics[width=0.9\linewidth]{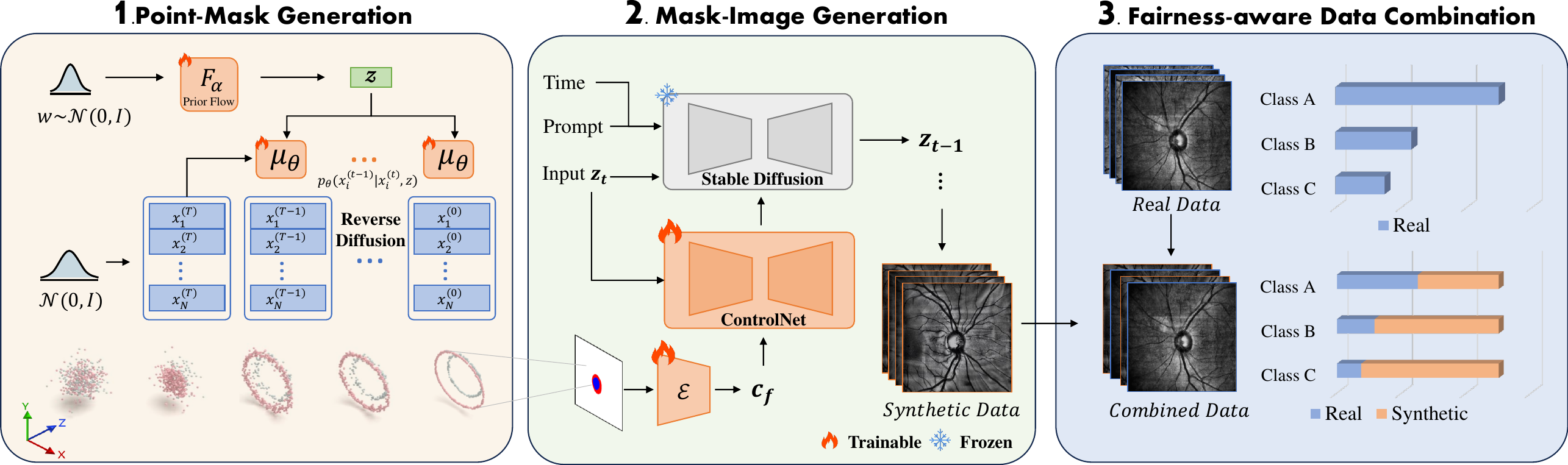}
    \caption{\textbf{Overview of Our Fairness-aware Point-Image Diffusion Framework.}}
    \label{fig:pipeline}
\end{figure}

\section{Methodology}
\subsection{Overview}

\textbf{Preliminaries.}
Fairness in a segmentation model can be defined as the model's ability to score evenly for images of different target groups. 
Consider a dataset $D$ that is made up of image pairs $(x, y)$. Here, $x$ represents the input image and $y$ is the ground truth segmentation mask. $\hat{y}$ refers to the predicted mask. 
To measure the fairness of segmentation, we introduce a metric called $\text{Fairness}_\theta(D)$ on the dataset $D$ of segmentation method $\theta$.
If we have sensitive attributes $S=\{s_0,...,s_i,...s_k\}$, each attribute can divide the population into groups $G=\{g_1,...,g_n\}$. The fairness metric for segmentation can be represented as:
\begin{equation}
    \text{Fairness}_\theta(D) = - \sum_{i=1}^{k} \left( \text{Var}_{G} \left[ \text{M}_\theta(\hat{y}, y | s_i) \right] \right)
\end{equation}
where $\text{M}_\theta(\hat{y}, y | s_i)$ is the performance metrics such as mIoU or Dice coefficient, $\text{Var}_{G}$ is the variance of metrics within the groups under $s_i$. 

\textbf{Overview of Framework.}
To achieve fairness, our research primarily explores data-driven methodologies, focusing on the use of synthetic data. As illustrated in Fig~\ref{fig:pipeline}, we introduce a novel Point-Image architecture designed specifically for the high-quality SLO fundus images synthesis. The first step is to generate segmentation masks from a Gaussian noise $\mathcal{N}$ using a 3D diffusion model with parameters $\phi$ , and then to generate images $x$ from the masks using a 2D diffusion model with parameters $\psi$. After integrating data from real datasets with synthesized data, we proceed to train them using a segmentation model that is parameterized by $\theta$. The overall optimization can be defined as:

\begin{equation}
    \min_{\theta} \max_{\phi, \psi} \text{Fairness}(D, \theta, \phi, \psi) 
\end{equation}

\subsection{Point-Mask Generation} 
To produce diverse cup-disc shape fundus images for SLO and acquire precise label maps for synthetic image data, we first augment segmentation masks using the labels from actual real-world data.

\textbf{Transformation to 3D Point Cloud.} Given a 2D mask image of size $W \times H$, where $W$ and $H$ are the width and height of the image. The function $f: I \rightarrow P$ maps $I$ to a 3D point cloud $P$ for training. $f$ is defined as follows:

\begin{equation}
f(I) = \{p_i = (x_i, y_i, z_i) \,|\, (x_i, y_i) \in \text{Boundary}(I), \, z_i = g((x_i, y_i), I)\}
\end{equation}
where $(x_i, y_i)$ are the coordinates of a pixel in $I$, $\text{Boundary}(I)$ represents those pixels that are situated at the segmentation boundaries, $g((x_i, y_i), I)$ is a function that assigns the $z_i$ value based on the pixel's position. $g$ is defined as follows:

\begin{equation}
g((x_i, y_i), I) = 
\begin{cases} 
z_0 & \text{if $(x_i, y_i)$ is on the boundary of the Cup} \\
-z_0 & \text{if $(x_i, y_i)$ is on the boundary of the Disc}
\end{cases}
\end{equation}

\textbf{Point Cloud Diffusion for Generation.} After converting the existing 2D labels into 3D point clouds, we employ a point cloud diffusion model based on \cite{luo2021diffusion} to learn the distribution of these point clouds.
The primary training goal of this model is to simulate the reverse of a random diffusion process, learning to move from a normal distribution to the distribution of real point clouds. During the training phase, we introduce varying degrees of random noise into the point clouds and ensure that the denoising model predicts noise that closely matches the actual noise added. 

For each group $g_i$ of the sensitive attribute $S$, we train a point cloud diffusion model $\phi_i$. Since $\phi_i$ can effectively capture the characteristics of the cup-disc contour for different groups, we can selectively augment samples for different groups, particularly for minority populations. Through this approach, we prepare the label sets for subsequent procedure.

\subsection{Mask-Image Generation} 

Given the generated mask $m$ as condition $c$, the next step is to synthesize an image $x$ . 
This integration of $m$ into the neural network is achieved by introducing an extra condition $c$ into a neural network block, via an architecture known as ControlNet \cite{zhang2023adding}.
This method involves freezing the original Stable Diffusion block's parameters $\Phi$, replicating it into a trainable copy with parameters $\Phi_c$, and connecting these blocks with two zero-initialized $1\times1$ convolutional layers. Specifically, the mask $m$ is encoded into tokens $c_f = E(c_i)$, which are then fed into ControlNet. The output of ControlNet $y_c$ is given by:
\begin{equation}
    y_c = F(x; \Phi) + Z\left(F\left(x + Z(c; \Phi_{z1}); \Phi_c\right); \Phi_{z2}\right)
\end{equation}
where $y_c$ is the output from the ControlNet block, $Z(\cdot; \cdot)$ represents the zero convolutional layers, and $\Phi_{z1}$ and $\Phi_{z2}$ are the parameters of the two zero convolutional layers. At the beginning of training, $y_c$ equals $y$ due to the zero initialization of the zero convolutional layers' weights and biases, ensuring no harmful noise is introduced into the network's hidden states. As training progresses, the zero convolutional layers gradually adapt the output based on the input condition $c_f$, thereby achieving control over the original feature map $x$.

\subsection{Equal-Scale Data Combination}
The method of combining real and synthetic data is straightforward, termed Equal-Scale Data Combination, which balances the sample sizes across all sensitive groups.
Assume $D_r = \{ x_{r,1}, x_{r,2}, \ldots, x_{r, N_r}\}$ and $D_s = \{x_{s,1}, x_{s,2}, \ldots, x_{s, N_s}\}$ as the sets of sample points from the real data distribution $P_r$ and the synthetic data distribution $P_s$, respectively. Here, $N_r$ and $N_s$ denote the number of samples in the real and synthetic datasets, respectively.
The equal-scale combination process involves augmenting the dataset with synthetic samples for underrepresented groups or possibly subsampling overrepresented groups. 
For each sensitive group $g$, if $N_{g,r} < N_{target}$, generate $(N_{target} - N_{g,r})$ synthetic samples from $P_s$ specific to group $g$, resulting in a combined set $D_{g,s}^{*}$. If $N_{g,r} > N_{target}$, random sample $N_{target}$ samples from $D_{g,r}^{*}$. $N_{target}$ is the target sample size, which could be based on the size of the largest group, a specified threshold for fairness. The combined dataset $D$ for training can be represented as:
\begin{equation}
    D = \bigcup_{g} (D_{g,r}^{*} \cup D_{g,s}^{*})
\end{equation}


\section{Experiments and Results}
\subsection{Setup}
\textbf{Datasets.} 
We use Harvard-FairSeg \cite{tian2024fairseg} as the real SLO fundus image dataset. It includes six critical attributes for comprehensive studies on fairness, which are age, gender, race, ethnicity, language preference, and marital status. The fairness and segmentation results of all models, whether using synthetic data or not, are tested on the test split of 2,000 real SLO fundus images. 

\textbf{Segmentation Models.}  To verify the impact of our synthetic data on segmentation and fairness, we selected two segmentation models, including a small model TransUNet \cite{chen2021TransUNet} and a larger model SAMed \cite{zhang2024segment} (the experiments on the latter are provided in the supplementary).

\textbf{Training Details.}
For the training of image synthesis, we utilize 512 points to sample the boundaries of each original mask. These point clouds are then normalized. Training is performed on NVIDIA 3090 GPUs with a batch size of 48 and a learning rate of 1e-4 across 2,000,000 steps. 
For the training of the segmentation model, following the experimental setup of Fairseg \cite{tian2024fairseg}, we employed a combination of cross-entropy and Dice losses as the training loss and used the AdamW optimizer. 
To enable effective comparisons, TransUNet was trained with a base learning rate of 0.01 and a momentum of 0.9 over 300 epochs, while SAMed was set with a base learning rate of 0.005 and a momentum of 0.9, undergoing training for 100 epochs. The batch size for both was set to 48. For the number of training samples, we have controlled it to be 8000, whether using all real data or a mix of real and synthetic data.

\subsection{Synthesis Quality results}
\textbf{Metrics.} To evaluate the generation quality, we employ the Fréchet inception distance (FID) \cite{heusel2017gans},
minimum matching distance (MMD) and the coverage score (COV). The detailed definitions of these metrics can be found in the \textit{supplementary materials}.

\textbf{Results.} We compare our Point-Image image generation method with several state-of-the-art  methods, including  Stabel Diffusion 1.5 \cite{rombach2022high}, pix2pixHD \cite{wang2018high}, OASIS \cite{sushko2020you}, SPADE \cite{park2019semantic} and ControlNet \cite{zhang2023adding}.
As shown in Tab.~\ref{tab:syn-quality}, our method significantly outperforms existing techniques in SLO fundus image synthesis.  Notably, our approach achieves the lowest FID score, indicating that our generated images bear a closer resemblance to the actual images when compared to other methods. Furthermore, the MMD results suggest that our method also more accurately replicates the distribution of the original image dataset. 

\begin{figure}[!t]
    \centering
    \includegraphics[width=0.9\linewidth]{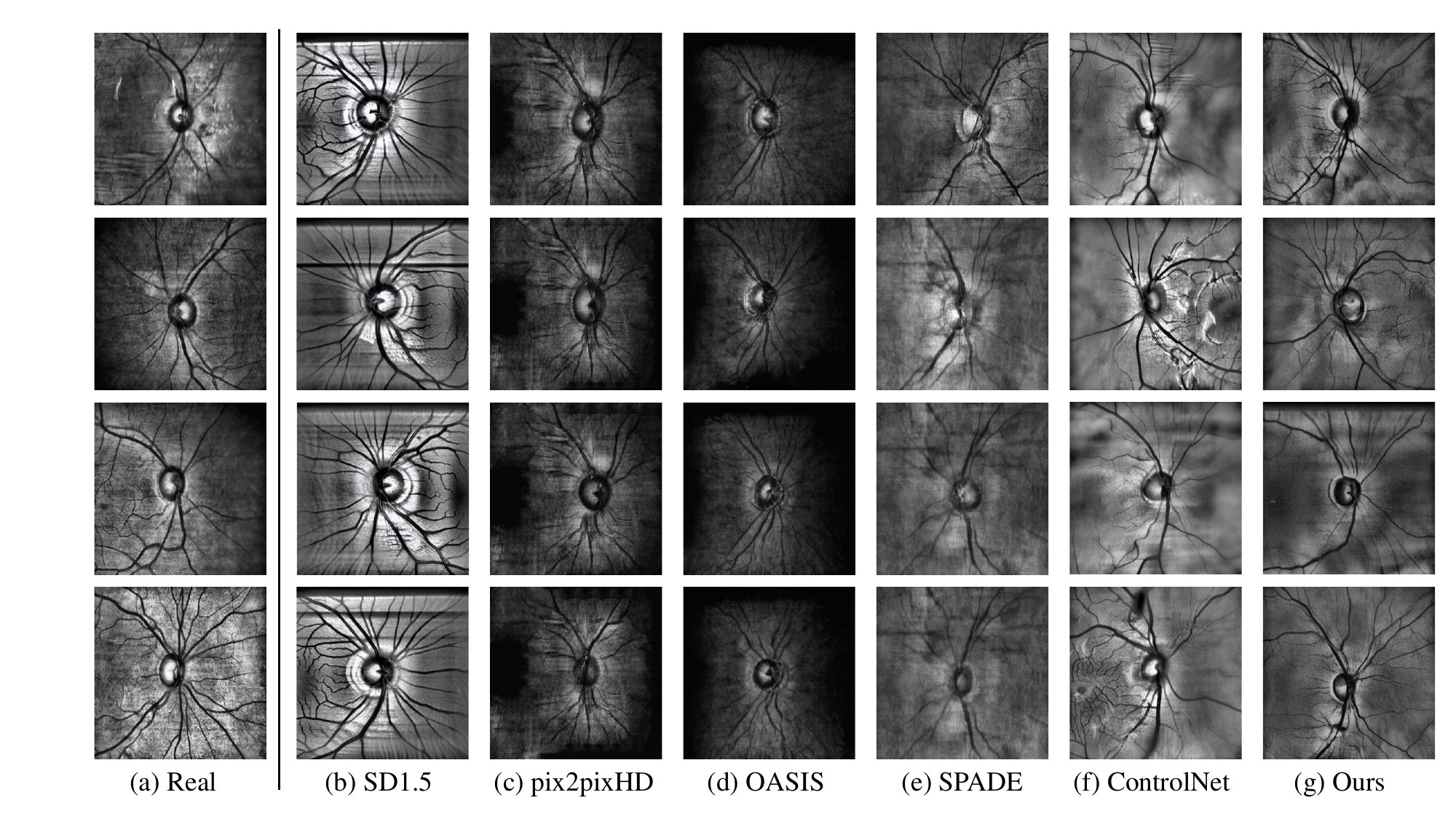}
    \caption{ \textbf{Visualization Results of Different Image Synthesis Results.}}
    \label{fig:compare}
\end{figure}

\textbf{Ablation over two-stage diffusion.} Comparing with ControlNet \cite{zhang2023adding} (one-stage label-to-image synthesis), our two-stage pipeline, where we first sample labels and then synthesis images, shows effectiveness in generating diverse images, as reflected by the highest COV (Coverage) score among the methods evaluated. The enhancement in image quality and diversity underscores the efficacy of our image synthesis technique. Fig.~\ref{fig:compare} visualizes the results of image synthesis.

\input{Table/synthesis_quality}

\subsection{Fairness Segmentation Results}
\textbf{Metrics.} Following prior work \cite{tian2024fairseg}, we measure the fairness segmentation results using Equity-Scaled Segmentation Performance (ESSP), which is defined as
\begin{equation}
    \text{ESSP} = \frac{\mathcal{L}(\{ (z', y) \})}{1 + \text{Stdev}}
\end{equation}
where $\mathcal{L}$ is the Dice or IoU metric. The ES-Dice and ES-IoU metrics consider both segmentation performance and fairness across all groups. The conventional Overall Dice and IoU metrics only assess the segmentation performance.

\input{Table/TransUNet_race}
\input{Table/TransUNet_gender}
\input{Table/TransUNet_language}
\input{Table/TransUNet_ethnicity}

\textbf{Results.} 
In our comparative analysis, we examine the performance of our Equal Scale Data Combination method against several state-of-the-art fairness-learning approaches, including ADV \cite{madras2018learning}, GroupDRO \cite{sagawa2019distributionally}, and FairSeg \cite{tian2024fairseg}. This evaluation encompasses experiments conducted across four sensitive attributes. The detailed results for TransUNet are presented from Table~\ref{tab:transunet-race} to Table~\ref{tab:TransUNet-ethnicity}. Due to limitations in space, the results of SAMed are included in the supplementary materials. From the perspective of racial fairness, Tab.~\ref{tab:transunet-race} highlights our Equal Scale method's effectiveness, achieving the highest ES-Dice in both the Cup and Rim area among all racial groups Asian, Black, and White, with remarkable scores of 0.8397 and 0.7697, respectively. The ES-IoU metric also supports our method, highlighting its effectiveness in achieving both accurate and equitable segmentation.
Tab.~\ref{tab:TransUNet-gender}, Tab.~\ref{tab:transunet-language} and Tab.~\ref{tab:TransUNet-ethnicity} also demonstrate the capability to enhance fairness metrics (ES-Dice \& ES-IoU) and segmentation performance (Dice \& IoU).

\section{Conclusion}
In this study, we analyze fairness within the context of medical image segmentation and address the challenge of data imbalance through the use of synthetic data. We present a novel Point-Image Diffusion method tailored for synthesizing SLO fundus images, which significantly outperforms existing techniques in this domain. By incorporating both synthetic and real data during the training phase utilizing the Equal Scale method, we achieve a comprehensive improvement in both accuracy and fairness across various sensitive attributes.


\newpage
\bibliographystyle{splncs04}

\input{main.bbl}
\newpage
\appendix

\input{appendix}

\end{document}

%% file: Table/synthesis_quality.tex
\begin{table}
\caption{\textbf{Comparison of Synthesis Quality.} }
\label{tab:syn-quality}
\centering
\setlength{\tabcolsep}{4pt}
\setlength{\extrarowheight}{0pt} 
\begin{tblr}{ccccc}
\toprule
Method    & FID↓   & MMD↓  & COV↑  &  \\ 
\midrule
SD1.5\cite{rombach2022high}   & 167.39 & 33.21 & 3.13  &  \\
pix2pixHD\cite{wang2018high} & 157.02 & 22.73 & 4.63  &  \\
OASIS\cite{sushko2020you}     & 89.92  & 28.57 & 3.07  &  \\
SPADE\cite{park2019semantic}     & 77.26  & 23.82 & 5.75  &  \\
\hline
ControlNet\cite{zhang2023adding}  (w/o Point-Mask)    & 67.29  & 23.60 & 9.45  &  \\
\textbf{Ours\textsuperscript}  (w/ Point-Mask)      & \textbf{60.51}  & \textbf{20.06} & \textbf{10.83} &  \\ 
\bottomrule
\end{tblr}
\vspace{-15pt}
\end{table}

%% file: Table/transunet_race.tex
\begin{table}[!ht]
\centering
\setlength{\tabcolsep}{8pt} 
\setlength{\extrarowheight}{2pt}
\caption{TransUNet segmentation performance of Optic Cup and Rim \textbf{(Sensitive attribute = Race)}}
\label{tab:transunet-race}
\adjustbox{max width=\textwidth}{
\begin{tabular}{clcccccccccc}
\hline
\multicolumn{1}{l}{} &                   & Overall & Overall & Overall & Overall & Asian  & Black  & White  & Asian  & Black  & White  \\
\multicolumn{1}{l}{} & Method            & ES-Dice$\uparrow$ & Dice$\uparrow$    & ES-IoU$\uparrow$  & IoU$\uparrow$     & Dice$\uparrow$   & Dice$\uparrow$   & Dice$\uparrow$   & IoU$\uparrow$    & IoU$\uparrow$    & IoU$\uparrow$    \\ \hline
\multirow{5}{*}{\rotatebox[origin=c]{90}{Cup}} 
                     & TransUNet             & 0.8372 & \textbf{0.8481} & 0.7409 & \textbf{0.7532} & 0.8270 & \textbf{0.8489} & \textbf{0.8503} & 0.7277 & \textbf{0.7576} & \textbf{0.7551} \\
                     & TransUNet+ADV      & 0.8325 & 0.8410 & 0.7345 & 0.7432 & 0.8246 & 0.8417 & 0.8426 & 0.7260 & 0.7482 & 0.7440 \\
                     & TransUNet+GroupDRO       & 0.8313 & 0.8442 & 0.7359 & 0.7479 & 0.8197 & 0.8469 & 0.8464 & 0.7232 & 0.7529 & 0.7495 \\
                    & TransUNet+FairSeg              & 0.8350 & 0.8464 & 0.7374 & 0.7497 & 0.8248 & 0.8484 & 0.8484 & 0.7247 & 0.7550 & 0.7513 \\
                     & \textbf{Ours(Equal Scale)}  & \textbf{0.8397} & 0.8480 & \textbf{0.7441} & 0.7529 & \textbf{0.8320} & 0.8483 & 0.8497 & \textbf{0.7352} & 0.7572 & 0.7540 \\ \hline
\multirow{5}{*}{\rotatebox[origin=c]{90}{Rim}} 
                     & TransUNet                 & 0.7604 & 0.7927 & 0.6393 & 0.6706 & 0.7457 & 0.7307 & 0.8106 & 0.6160 & 0.5991 & 0.6913 \\
                     
                     & TransUNet+ADV       & 0.7579 & 0.7906 & 0.6371 & 0.6682 & 0.7413 & 0.7286 & 0.8087 & 0.6116 & 0.5982 & 0.6888 \\
                     & TransUNet+GroupDRO        & 0.7564 & 0.7896 & 0.6351 & 0.6674 & 0.7470 & 0.7229 & 0.8080 & 0.6183 & 0.5899 & 0.6887 \\
                     & TransUNet+FairSeg              & 0.7628 & 0.7950 & 0.6410 & 0.6725 & 0.7479 & 0.7325 & 0.8130 & 0.6185 & 0.6020 & 0.6935 \\
                     & \textbf{Ours(Equal Scale)}  & \textbf{0.7697} & \textbf{0.7999} & \textbf{0.6494} & \textbf{0.6797} & \textbf{0.7565} & \textbf{0.7427} & \textbf{0.8165} & \textbf{0.6279} & \textbf{0.6114} & \textbf{0.6994} \\ \hline

\end{tabular}}
\end{table}

%% file: Table/transunet_gender.tex
\begin{table}[!ht]
\centering
\setlength{\tabcolsep}{8pt} 
\setlength{\extrarowheight}{2pt}
\caption{TransUNet segmentation performance of Optic Cup and Rim \textbf{(Sensitive attribute = Gender)}}
\label{tab:TransUNet-gender}
\adjustbox{max width=\textwidth}{
\begin{tabular}{clcccccccccc}
\hline
\multicolumn{1}{l}{} &                   & Overall & Overall & Overall & Overall & Male  & Female    & Male  & Female    \\
\multicolumn{1}{l}{} & Method            & ES-Dice$\uparrow$ & Dice$\uparrow$    & ES-IoU$\uparrow$  & IoU$\uparrow$     & Dice$\uparrow$   & Dice$\uparrow$      & IoU$\uparrow$    & IoU$\uparrow$        \\ \hline
\multirow{5}{*}{\rotatebox[origin=c]{90}{Cup}} 
                      & TransUNet                  & 0.8448       & 0.8481       & 0.7502       & 0.7532       & 0.8458        & 0.8513       & 0.7508        & 0.7564        \\
                     & TransUNet+ADV             & 0.8343       & 0.8345       & 0.7351       & 0.7356       & 0.8344        & 0.8348       & 0.7361        & 0.7350        \\
                     & TransUNet+GroupDRO        & 0.8426       & 0.8478       & 0.7473       & 0.7522       & 0.8441        & 0.8528       & 0.7483        & 0.7575        \\
                     & TransUNet+FairSeg                   &\textbf{ 0.8477}       & 0.8489       & 0.7502       & 0.7530       & \textbf{0.8494}        & 0.8514       & 0.7505        & 0.7556        \\
                     & \textbf{Ours(Equal Scale)}          & 0.8461     & \textbf{0.8505}       & \textbf{0.7522}       & \textbf{0.7564}       & 0.8474        & \textbf{0.8548}      & \textbf{0.7531}        & \textbf{0.7610}        \\ \hline
\multirow{5}{*}{\rotatebox[origin=c]{90}{Rim}} 
                     & TransUNet                  & 0.7895       & 0.7927       & 0.6673       & 0.6706       & 0.7951        & 0.7894       & 0.6736        & 0.6665        \\
                     & TransUNet+ADV             & 0.7783       & 0.7852       & 0.6553       & 0.6630       & 0.7905        & 0.7779       & 0.6699        & 0.6534        \\
                     & TransUNet+GroupDRO        & 0.7901       & 0.7917       & 0.6681       & 0.6699       & 0.7930        & 0.7900       & 0.6716        & 0.6677        \\
                     & TransUNet+FairSeg                   & 0.7893       & 0.7898       & 0.6698       & 0.6698       & 0.7924        & 0.7932       & 0.6678        & 0.6653        \\
                     & \textbf{Ours(Equal Scale)}          & \textbf{0.7945}       & \textbf{0.7981}       & \textbf{0.6745}      & \textbf{0.6780}       & \textbf{0.8007}        &\textbf{ 0.7944}       & \textbf{0.6811}        & \textbf{0.6737}        \\ \hline
\end{tabular}}
\end{table}

%% file: Table/transunet_language.tex
\begin{table}[!ht]
\centering
\setlength{\tabcolsep}{5pt} 
\setlength{\extrarowheight}{2pt}
\caption{TransUNet segmentation performance of Optic Cup and Rim \textbf{(Sensitive attribute = Language)}}
\label{tab:transunet-language}
\adjustbox{max width=\textwidth}{
\begin{tabular}{clcccccccccc}
\hline
\multicolumn{1}{l}{} &                   & Overall & Overall & Overall & Overall & English  & Spanish  & Others  & English  & Spanish  & Others  \\
\multicolumn{1}{l}{} & Method            & ES-Dice$\uparrow$ & Dice$\uparrow$    & ES-IoU$\uparrow$  & IoU$\uparrow$     & Dice$\uparrow$   & Dice$\uparrow$   & Dice$\uparrow$   & IoU$\uparrow$    & IoU$\uparrow$    & IoU$\uparrow$    \\ \hline
\multirow{5}{*}{\rotatebox[origin=c]{90}{Cup}} 
                     & TransUNet                 & 0.8255 & 0.8481 & 0.7273 & 0.7532 & 0.8469 & 0.8972 & 0.8531 & 0.7516 & \textbf{0.8166} & 0.7592 \\
                     & TransUNet+ADV       & 0.8071 & 0.8312 & 0.7056 & 0.7323 & 0.8296 & 0.8833 & 0.8338 & 0.7301 & 0.7990 & 0.7376 \\
                     & TransUNet+GroupDRO        & 0.8231 & 0.8416 & 0.7231 & 0.7442 & 0.8398 & 0.8844 & 0.8571 & 0.7421 & 0.7993 & 0.7605 \\
                     & TransUNet+FairSeg              & 0.8277 & 0.8481 & 0.7289 & 0.7523 & 0.8467 & \textbf{0.8934} & 0.8562 & 0.7504 & 0.8109 & 0.7619 \\
                     & \textbf{Ours(Equal Scale)}  & \textbf{0.8358} & \textbf{0.8497} & \textbf{0.7353} & \textbf{0.7542} & \textbf{0.8479} & 0.8809 & \textbf{0.8686} & \textbf{0.7519} & 0.8033 & \textbf{0.7755} \\ \hline
\multirow{5}{*}{\rotatebox[origin=c]{90}{Rim}} 
                     & TransUNet                 & 0.7721 & 0.7927 & 0.6525 & 0.6706 & 0.7940 & \textbf{0.8165} & 0.7633 & 0.6721 & 0.6950 & 0.6398 \\                     
                     & TransUNet+ADV       & 0.7690 & 0.7884 & 0.6501 & 0.6666 & 0.7903 & 0.7964 & 0.7501 & 0.6687 & 0.6717 & 0.6265 \\
                     & TransUNet+GroupDRO        & 0.7691 & 0.7857 & 0.6468 & 0.6613 & 0.7867 & 0.8057 & 0.7628 & 0.6625 & 0.6800 & 0.6355 \\
                     & TransUNet+FairSeg              & 0.7725 & 0.7898 & 0.6524 & 0.6668 & 0.7909 & 0.8106 & 0.7661 & 0.6680 & 0.6865 & 0.6424 \\
                     & \textbf{Ours(Equal Scale)}  & \textbf{0.7783} & \textbf{0.7959} & \textbf{0.6578} & \textbf{0.6743} & \textbf{0.7970}  & 0.8161 & \textbf{0.7711} & \textbf{0.6755} & \textbf{0.6969} & \textbf{0.6469} \\ \hline

\end{tabular}}
\end{table}

%% file: Table/transunet_ethnicity.tex
\begin{table}[!ht]
\centering
\setlength{\tabcolsep}{6pt} 
\setlength{\extrarowheight}{2pt}
\caption{TransUNet segmentation performance of Optic Cup and Rim \textbf{(Sensitive attribute = Ethnicity)}}
\label{tab:TransUNet-ethnicity}
\adjustbox{max width=\textwidth}{
\begin{tabular}{clcccccccccc}
\hline
\multicolumn{1}{l}{} &                   & Overall & Overall & Overall & Overall & Hispanic  & Non-Hispanic    & Hispanic  & Non-Hispanic    \\
\multicolumn{1}{l}{} & Method            & ES-Dice$\uparrow$ & Dice$\uparrow$    & ES-IoU$\uparrow$  & IoU$\uparrow$     & Dice$\uparrow$   & Dice$\uparrow$      & IoU$\uparrow$    & IoU$\uparrow$        \\ \hline
\multirow{5}{*}{\rotatebox[origin=c]{90}{Cup}} 
                     & TransUNet                 & 0.8339       & 0.8481       & 0.7366       & 0.7532       & 0.8463        & \textbf{0.8704}       & 0.7508        & \textbf{0.7826}        \\
                     & TransUNet+ADV       & 0.8171       & 0.8320       & 0.7149       & 0.7315       & 0.8304        & 0.8561       & 0.7294        & 0.7622        \\
                     & TransUNet+GroupDRO        & 0.8376       & 0.8482       & 0.7406       & 0.7526       & 0.8468        & 0.8648       & 0.7507        & 0.7735        \\
                     & TransUNet+FairSeg                   & 0.8388       & \textbf{0.8483}       & \textbf{0.7412}       & \textbf{0.7542}       & \textbf{0.8501}        & 0.8661       & 0.7515        & 0.7764        \\
                     & \textbf{Ours(Equal Scale)}  & \textbf{0.8439}       & 0.8462       & 0.7402       & 0.7500       & 0.8485        & 0.8448       & \textbf{0.7664}        & 0.7477        \\ \hline
\multirow{5}{*}{\rotatebox[origin=c]{90}{Rim}} 
                     & TransUNet                 & 0.7848       & 0.7927       & 0.6650       & 0.6706       & 0.7914        & \textbf{0.8057}       & 0.6695        & \textbf{0.6815}        \\
                     & TransUNet+ADV       & 0.7793       & 0.7841       & 0.6570       & 0.6602       & 0.7829        & 0.7915       & 0.6590        & 0.6658        \\
                     & TransUNet+GroupDRO        & 0.7924       & 0.7943       & 0.6694       & \textbf{0.6733}       & 0.7936        & 0.7901       & 0.6728        & 0.6646        \\
                     & TransUNet+FairSeg                   & 0.7884       & \textbf{0.7939}       & \textbf{0.6710}       & 0.6754       & \textbf{0.7943}        & 0.8040       & \textbf{0.6697}        & 0.6789        \\
                     & \textbf{Ours(Equal Scale)} & \textbf{0.7886}       & 0.7902       & 0.6666       & 0.6670       & 0.7917        & 0.7888       & 0.6649        & 0.6657        \\ \hline
\end{tabular}}
\end{table}

%% file: appendix.tex
\section{Point Cloud Diffusion for Generation} 

The forward diffusion process can be modeled as a Markov chain.

\begin{equation}
    q(x^{(1:T)}_i | x^{(0)}_i) = \prod_{t=1}^{T} q(x^{(t)}_i | x^{(t-1)}_i)
\end{equation}

The training process is helping the model to learn the flow from original shape distribution to a noise distribution, and learn the noise predictor $\theta$ of each step.The generation of point clouds can be treated as the reverse of the diffusion process.

\begin{equation}
p_{\theta}(x^{(0:T)}|z) = p(x^{(T)}) \prod_{t=1}^{T} p_{\theta}(x^{(t-1)}|x^{(t)}, z)
\end{equation}
\begin{equation}
    p_{\theta}(x^{(t-1)}|x^{(t)}, z) = \mathcal{N}(x^{(t-1)} | \mu_{\theta}(x^{(t)}, t, z), \beta_t I)
\end{equation}
where $\mu_\theta$ is the estimated mean implemented by a neural network parameterized by $\theta$. $z$ is the latent encoding the target shape of the point cloud. The starting distribution $ p(\mathbf{x}^{(T)}) $ is set to a standard normal distribution $ \mathcal{N}(0, \mathbf{I}) $.

~\\




\section{Synthesis Quality Metrics} 

The MMD-score measures the fidelity of generated samples, calculates the mean of the minimum matching distances between generated samples and real samples, used to evaluate the quality of the generative model. We define the distance $D$ between image $I_1$ and image $I_2$ as 
\begin{equation}
    D(I_1, I_2) = \frac{1 - \cos(\theta)}{2}
\end{equation}
where $\cos(\theta)$ represents the cosine similarity between the two images.
The COV-score denotes the proportion of real samples that match at least one image in the generated images, for generated set $S_g$ and the reference real set $S_r$, the COV-score is
\begin{equation}
    \text{COV}(S_g, S_r) =\frac{|\{\arg\min_{I_2 \in S_r} D(I_1, I_2) | I_1 \in S_g \}|}{|S_r|}
\end{equation}

\section{More Experimental Results} 

\input{Table/SAMed_race}
\input{Table/SAMed_gender}
\input{Table/SAMed_language}
\input{Table/SAMed_ethnicity}

%


%% file: Table/samed_race.tex
\begin{table}[!ht]
\centering
\setlength{\tabcolsep}{8pt} 
\setlength{\extrarowheight}{2pt}
\caption{SAMed segmentation performance of Optic Cup and Rim \textbf{(Sensitive attribute = Race)}}
\label{tab:samed-race}
\adjustbox{max width=\textwidth}{
\begin{tabular}{clcccccccccc}
\hline
\multicolumn{1}{l}{} &                   & Overall & Overall & Overall & Overall & Asian  & Black  & White  & Asian  & Black  & White  \\
\multicolumn{1}{l}{} & Method            & ES-Dice$\uparrow$ & Dice$\uparrow$    & ES-IoU$\uparrow$  & IoU$\uparrow$     & Dice$\uparrow$   & Dice$\uparrow$   & Dice$\uparrow$   & IoU$\uparrow$    & IoU$\uparrow$    & IoU$\uparrow$    \\ \hline
\multirow{5}{*}{\rotatebox[origin=c]{90}{Cup}} 
                     & SAMed            & 0.8600  & 0.8671  & 0.7729  & 0.7813  & 0.8568 & \textbf{0.8730} & 0.8670 & 0.7688 & \textbf{0.7905} & 0.7808 \\
                     & SAMed+ADV           & 0.8640  & \textbf{0.8698 } & \textbf{0.7769}  & \textbf{0.7840}  & 0.8590 & 0.8705 & 0.8708 & 0.7709 & 0.7882 & 0.7846 \\
                     & SAMed+GroupDRO       & \textbf{0.8634}  & 0.8695  & 0.7767  & 0.7838  & 0.8583 & 0.8704 & 0.8706 & 0.7711 & 0.7886 & \textbf{0.7842} \\
                     & SAMed+FairSeg       & 0.8617  & 0.8671  & 0.7741  & 0.7808  & 0.8587 & 0.8708 & \textbf{0.8672} & 0.7708 & 0.7882 & 0.7804 \\
                     & \textbf{Ours}         & 0.8619  & 0.8660  & 0.7737  & 0.7796  & \textbf{0.8606} & 0.8702 & 0.8657 & \textbf{0.7744} & 0.7892 & 0.7782 \\ \hline
\multirow{5}{*}{\rotatebox[origin=c]{90}{Rim}} 
                     & SAMed                  & 0.8000  & 0.8291  & 0.6919  & 0.7217  & 0.7890 & 0.7758 & 0.8444 & 0.6743 & 0.6587 & 0.7399 \\
                     & SAMed+ADV              & 0.7935  & 0.8235  & 0.6835  & 0.7138  & 0.7801 & 0.7691 & 0.8395 & 0.6635 & 0.6498 & 0.7325 \\
                     & SAMed+GroupDRO           & 0.8011  & 0.8302  & 0.6930  & 0.7230  & 0.7952 & 0.7748 & \textbf{0.8454} & 0.6822 & 0.6568 & 0.7410 \\
                     & SAMed+FairSeg            & 0.8036  & 0.8323  & 0.6963  & 0.7260  & 0.7952 & 0.7789 & 0.8473 & 0.6825 & 0.6620 & \textbf{0.7439} \\
                     & \textbf{Ours}      & \textbf{0.8041}  & \textbf{0.8311}  & \textbf{0.6966}  & \textbf{0.7242}  & \textbf{0.7968} & \textbf{0.7808} & 0.8452 & \textbf{0.6840} & \textbf{0.6646} & 0.7409 \\ \hline
\end{tabular}}
\end{table}

%% file: Table/samed_gender.tex
\begin{table}[!ht]
\centering
\setlength{\tabcolsep}{8pt} 
\setlength{\extrarowheight}{2pt}
\caption{SAMed segmentation performance of Optic Cup and Rim \textbf{(Sensitive attribute = Gender)}}
\label{tab:samed-gender}
\adjustbox{max width=\textwidth}{
\begin{tabular}{clcccccccccc}
\hline
\multicolumn{1}{l}{} &                   & Overall & Overall & Overall & Overall & Male  & Female   & Male  & Female   \\
\multicolumn{1}{l}{} & Method            & ES-Dice$\uparrow$ & Dice$\uparrow$    & ES-IoU$\uparrow$  & IoU$\uparrow$     & Dice$\uparrow$   & Dice$\uparrow$    & IoU$\uparrow$    & IoU$\uparrow$       \\ \hline
\multirow{5}{*}{\rotatebox[origin=c]{90}{Cup}} 
                      & SAMed                  & 0.8637       & 0.8671       & 0.7773       & 0.7813       & 0.8647       & 0.8703       & 0.7783       & 0.7855       \\
                     & SAMed+ADV              & 0.8658       & 0.8667       & 0.7787       & 0.7803       & 0.8661       & 0.8675       & 0.7791       & 0.7820       \\
                     & SAMed+GroupDRO         & 0.8670       & 0.8671       & 0.7803       & 0.7808       & 0.8672       & 0.8670       & 0.7804       & 0.7814       \\
                     & SAMed+FairSeg                & \textbf{0.8678}       & \textbf{0.8702}       & 0.7807       & 0.7823       & \textbf{0.8718}       & \textbf{0.8756}       & \textbf{0.7851}       & 0.7879       \\
                     & \textbf{Ours}      & 0.8676       & 0.8698       & \textbf{0.7809}       & \textbf{0.7844}       & 0.8683       & 0.8718       & 0.7817       & \textbf{0.7881}       \\ \hline
\multirow{5}{*}{\rotatebox[origin=c]{90}{Rim}} 
                     & SAMed                  & 0.8251       & 0.8291       & 0.7175       & 0.7217       & 0.8319       & 0.8252       & 0.7252       & 0.7169       \\
                     & SAMed+ADV              & 0.8263       & 0.8309       & 0.7188       & 0.7236       & 0.8342       & 0.8263       & 0.7276       & 0.7181       \\
                     & SAMed+GroupDRO         & 0.8274       & 0.8320       & 0.7205       & 0.7253       & 0.8353       & 0.8274       & 0.7292       & 0.7198       \\
                     & SAMed+FairSeg                & \textbf{0.8289}       & \textbf{0.8318}       & \textbf{0.7227}       & \textbf{0.7253}       & \textbf{0.8338}       & \textbf{0.8289}       & \textbf{0.7274}       & \textbf{0.7223}       \\
                     & \textbf{Ours}      & 0.8221       & 0.8265       & 0.7132       & 0.7177       & 0.8297       & 0.8221       & 0.7214       & 0.7125       \\ \hline
\end{tabular}}
\end{table}

%% file: Table/samed_language.tex
\begin{table}[!ht]
\centering
\setlength{\tabcolsep}{8pt} 
\setlength{\extrarowheight}{2pt}
\caption{SAMed segmentation performance of Optic Cup and Rim \textbf{(Sensitive attribute = Language)}}
\label{tab:samed-language}
\adjustbox{max width=\textwidth}{
\begin{tabular}{clcccccccccc}
\hline
\multicolumn{1}{l}{} &                   & Overall & Overall & Overall & Overall & English  & Spanish  & Others  & English  & Spanish  & Others  \\
\multicolumn{1}{l}{} & Method            & ES-Dice$\uparrow$ & Dice$\uparrow$    & ES-IoU$\uparrow$  & IoU$\uparrow$     & Dice$\uparrow$   & Dice$\uparrow$   & Dice$\uparrow$   & IoU$\uparrow$    & IoU$\uparrow$    & IoU$\uparrow$    \\ \hline
\multirow{5}{*}{\rotatebox[origin=c]{90}{Cup}} 
                     & SAMed                 & 0.8490  & 0.8671  & 0.7603  & 0.7813  & 0.8652 & 0.9077 & 0.8838 & 0.7791 & 0.8338 & 0.8001 \\
                     & SAMed+ADV   & 0.8485 & 0.8686 & 0.7586 & 0.7830 & 0.8668 & \textbf{0.9131} & 0.8820 & 0.7808 & \textbf{0.8432} & 0.7982  \\
                     & SAMed+GroupDRO  &0.8530 & \textbf{0.8702} & 0.7640 & \textbf{0.7847} & \textbf{0.8684} & 0.9085 & \textbf{0.8849} & \textbf{0.7825} & 0.8360 & \textbf{0.8019} \\
                     & SAMed+FairSeg    & 0.8527  & 0.8684  & \textbf{0.7646}  & 0.7826  & 0.8670 & 0.9034 & 0.8794 & 0.7810 & 0.8268 & 0.7937 \\
                     & \textbf{Ours}        & \textbf{0.8518}  & 0.8676  & 0.7624  & 0.7810  & 0.8659 & 0.9029 & 0.8815 & 0.7789 & 0.8271 & 0.7968 \\   \hline
\multirow{5}{*}{\rotatebox[origin=c]{90}{Rim}} 
                     & SAMed                     & 0.8070  & 0.8291  & 0.7006  & 0.7217  & 0.8305 & \textbf{0.8534} & 0.7989 & 0.7234 & \textbf{0.7468} & 0.6871 \\
                     & SAMed+ADV   & 0.8087 & 0.8295 & 0.7019 & 0.7217 & 0.8307 & 0.8528 & 0.8015 & 0.7231 & 0.7463 & 0.6900 \\
                     & SAMed+GroupDRO           &\textbf{0.8136} & 0.8311 & \textbf{0.7075} & 0.7239 & 0.8322 & 0.8493 & \textbf{0.8065} & 0.7253 & 0.7411 & \textbf{0.6954} \\
                     & SAMed+FairSeg            & 0.8100  & \textbf{0.8313}  & 0.7038  & \textbf{0.7244} & \textbf{0.8328} & 0.8511 & 0.7992 & \textbf{0.7263} & 0.7436 & 0.6865 \\
                     & \textbf{Ours}      & 0.8036  & 0.8245  & 0.6944  & 0.7145  & 0.8258 & 0.8472 & 0.7955 & 0.7160 & 0.7377 & 0.6805 \\ \hline
\end{tabular}}
\end{table}

%% file: Table/samed_ethnicity.tex
\begin{table}[!ht]
\centering
\setlength{\tabcolsep}{8pt} 
\setlength{\extrarowheight}{2pt}
\caption{SAMed segmentation performance of Optic Cup and Rim \textbf{(Sensitive attribute = Ethnicity)}}
\label{tab:samed-ethnicity}
\adjustbox{max width=\textwidth}{
\begin{tabular}{clcccccccccc}
\hline
\multicolumn{1}{l}{} &                   & Overall & Overall & Overall & Overall & Hispanic  & Non-Hispanic   & Hispanic  & Non-Hispanic   \\
\multicolumn{1}{l}{} & Method            & ES-Dice$\uparrow$ & Dice$\uparrow$    & ES-IoU$\uparrow$  & IoU$\uparrow$     & Dice$\uparrow$   & Dice$\uparrow$    & IoU$\uparrow$    & IoU$\uparrow$       \\ \hline
\multirow{5}{*}{\rotatebox[origin=c]{90}{Cup}} 
                     & SAMed                 & 0.8519       & 0.8671       & 0.7645       & 0.7813       & 0.8653       & \textbf{0.8904 }      & 0.7790       & \textbf{0.8100}       \\
                     & SAMed+ADV       & 0.8544       & 0.8678       & 0.7657       & 0.7814       & 0.8661       & 0.8883       & 0.7791       & 0.8080       \\
                     & SAMed+GroupDRO        & 0.8594       & \textbf{0.8698}       & 0.7718       & 0.7840       & 0.8682       & 0.8855       & 0.7819       & 0.8044       \\
                     & SAMed+FairSeg                   & 0.8611       & 0.8685       & \textbf{0.7753}       &\textbf{ 0.7845}       & 0.8704       & 0.8824       & 0.7904       & 0.8070       \\
                     & \textbf{Ours}      & \textbf{0.8625}       & 0.8664       & 0.7730       & 0.7793       & \textbf{0.8714}       & 0.8650       & \textbf{0.7889}       & 0.7775       \\  \hline
\multirow{5}{*}{\rotatebox[origin=c]{90}{Rim}} 
                     & SAMed                 & 0.8221       & 0.8291       & 0.7164       & 0.7217       & 0.8277       & 0.8397       & 0.7203       & 0.7307       \\
                     & SAMed+ADV       & 0.8260       & 0.8323       & 0.7206       & \textbf{0.7257}       & 0.8308       & \textbf{0.8416}       & 0.7241       & \textbf{0.7342}       \\
                     & SAMed+GroupDRO        & 0.8237       & 0.8299       & 0.7178       & 0.7224       & 0.8284       & 0.8390       & 0.7208       & 0.7298       \\
                     & SAMed+FairSeg                   & \textbf{0.8296}       & \textbf{0.8331}       & \textbf{0.7215}       & 0.7242       & \textbf{0.8349}       & 0.8408       & \textbf{0.7278}       & 0.7329       \\
                     & \textbf{Ours}      & 0.8186       & 0.8234       & 0.7112       & 0.7136       & 0.8306       & 0.8222       & 0.7171       & 0.7124       \\  \hline
\end{tabular}}
\end{table}